# On Evaluation of Embodied Navigation Agents

Peter Anderson
Australian National University

Angel Chang
Princeton University

Devendra Singh Chaplot
Carnegie Mellon University

Alexey Dosovitskiy
Intel Labs

Saurabh Gupta
UC Berkeley

Vladlen Koltun
Intel Labs

Jana Kosecka
George Mason University

Jitendra Malik
UC Berkeley and Facebook

Roozbeh Mottaghi
Allen Institute for AI

Manolis Savva
Princeton University

Amir R. Zamir
Stanford and UC Berkeley

## Abstract

*Skillful mobile operation in three-dimensional environments is a primary topic of study in Artificial Intelligence. The past two years have seen a surge of creative work on navigation. This creative output has produced a plethora of sometimes incompatible task definitions and evaluation protocols. To coordinate ongoing and future research in this area, we have convened a working group to study empirical methodology in navigation research. The present document summarizes the consensus recommendations of this working group. We discuss different problem statements and the role of generalization, present evaluation measures, and provide standard scenarios that can be used for benchmarking.*

## 1. Introduction

Navigation in three-dimensional environments is an essential capability of mobile intelligent systems that function in the physical world. Animals, including humans, can traverse cluttered dynamic environments with grace and skill in pursuit of distal goals. Animals can navigate efficiently and deliberately in previously unseen environments, building up internal representations of these environments in the process. Such internal representations are of central importance to Artificial Intelligence.

Earlier work on navigation has been somewhat fragmented. There is a large body of work on motion planning, where a collision-free path in workspace or configuration space is sought, given a geometric model of the environment [18]. Subsequent path following and low-level control usually assume perfect localization. Motion planning approaches often rely on high-quality geometric models of the environment, limiting the use of these methods.

There is also a rich and informative body of work on Simultaneous Localization and Mapping (SLAM) [6], which focuses on building a map of the environment and localizing the agent within the map. Traditional SLAM systems largely focus on constructing metric maps using geometric techniques. Navigation itself (i.e., what to do, where to go) is rarely considered. Representations constructed by SLAM systems are often not suitable for traditional motion planning methods and are prone to error when the environment changes over time.

In contrast, biological systems appear to use more flexible representations, can robustly navigate without precise localization or a metric map, and can bring substantial prior knowledge to bear when planning in previously unseen environments.

We view skillful mobile operation in three-dimensional environments as a primary topic of study in Artificial Intelligence. We anticipate a renaissance of research on navigation, informed by diverse perspectives and ongoing advances in machine learning, perception, and reasoning. Indeed, the past two years have seen a surge of creative work in the area [1, 3, 12, 13, 19, 20, 22, 23, 27, 32].

This creative output comes with a diversity of task definitions, experimental testbeds, and evaluation protocols. While creativity in technique is desirable, a kaleidoscope of incompatible experimental methodologies can be detrimental. Given the importance of the area and the growing amount and diversity of related research, we believe there is a need to coordinate efforts via the common task framework [8]. To this end, we have convened a working group to study empirical methodology in navigation research. The present document summarizes the consensus recommendations of this working group.

Convergence to common task definitions and evaluation protocols catalyzed dramatic progress in computer vision [11, 25]. We hope that the presented recommendations will contribute to similar momentum in navigation research.

---

The authors are listed in alphabetical order.

## 2. Goal Specification and Sensory Input

Navigation tasks can be distinguished along several dimensions. One is the nature of the goal. We identify three types of goals:

- **PointGoal.** The agent must navigate to a specific location. For example, assuming the agent starts at the origin, the goal may be to navigate to location $(100, 300)$, where the units are in meters. This task is trivial if the environment is empty, but is beyond the capability of existing systems in realistic cluttered environments that have not been previously explored [28].

- **ObjectGoal.** The agent must navigate to an object of a specific category. The category can be drawn from a predefined set. For example, 'refrigerator', 'car', or 'keys'. To perform this task, the agent must draw on prior knowledge about the world, such as what a 'refrigerator' looks like and where it may be found.

- **AreaGoal.** The agent must navigate to an area of a specified category. For example, 'kitchen', 'garage', or 'foyer'. This task also relies on prior knowledge about the appearance and layout of different areas.

The different types of goals can be specified by different means. The basic specification is exemplified in the preceding descriptions: coordinates for PointGoal, categorical labels for ObjectGoal and AreaGoal. There are two other types of specifications that are interesting and noteworthy: images (or other perceptual input) and language. For example, an ObjectGoal task can be specified by an image of the object in question [32]. A PointGoal task can be specified by a description in natural language [1]. Some of these specification modalities also support an instance-specific form of ObjectGoal and AreaGoal tasks: for example, find the specific car shown in this image (rather than any car).

The agent may be equipped with different sensing modalities, such as vision (RGB image), depth, or haptics. Another possible input is a schematic map of the environment, of the kind used by humans when navigating. Some recent work also relies on idealized perception of self-motion (perfect odometry). Our primary recommendation in this regard is that the sensory input available to the agent be made explicit when presenting the method. For example, if the method assumes perfect odometry or GPS for localization, this must be stated explicitly.

## 3. Generalization and Exploration

Another principal distinction is the extent of prior exposure to the environment in which the agent is evaluated. Recent work spans a broad spectrum. On one end are protocols that evaluate agents in new environments, with no exposure to the environment prior to the test episode [12, 28]. On the other end are protocols that conduct extensive training in the same environment in which the agent is then tested, giving the agent days or weeks of training on the test scene [20]. Along the spectrum are protocols that give the agent brief exposure (on the order of minutes of subjective experience) to the test environment prior to evaluation; during this preliminary exposure the agent can build an internal representation that is then used to support navigation [27].

Our basic principle is that the extent of the agent's exposure to the test environment prior to the navigation episode should be rigorously quantified and reported. Approaches that require extensive experience in the test environment prior to navigation trials may be justified, but this prior exposure to the environment must be clearly quantified and documented. We can identify a number of regimes with respect to generalization:

- **No prior exploration.** No prior exposure to a test environment. Agents are asked to find their way around novel, previously unseen environments [12, 28].

- **Pre-recorded prior exploration.** The agent is given a recording of an exploration trajectory through the environment. Exploration was performed by a third party (e.g., a human or an automated exploration policy) and a recording is provided with each test environment, as part of the benchmarking setup. The agent can use the provided recording to construct an internal representation of the environment that can support subsequent navigation [27].

- **Time-limited exploration by the agent.** The agent is given a budget for exploration. Prior to a navigation episode, the agent is free to traverse the environment until the length of its trajectory reaches the given budget. This experience can be used to construct an internal representation of the environment that can support subsequent navigation episodes. The exploration policy is under the agent's control, but the extent of exposure to the test environment is still limited and quantified.

  In this regime, the exploration budget given to the agent can be varied, such that the navigation performance of the agent is evaluated as a function of exploration extent. (E.g., navigation performance after 500 meters of exploration, 1,000 meters, 2,000 meters, etc.) This can yield a multi-objective performance profile that quantifies the exploration-navigation trade-off for different agents. An agent can aim to dominate across the range of exploration budgets, or at least contribute to the Pareto front.

The role of exploration in navigation suggests that exploration is itself an important task that can be studied and quantitatively evaluated. For example, an agent can be tasked with exploring a previously unseen environment within a path-length budget (maximal trajectory length, at the end of which the exploration episode is terminated). One possible measure for evaluating an exploration episode is the coverage of the environment: what fraction of the previously unseen environment the agent has covered during the episode.

## 4. Evaluation Measures

Broad agreement on evaluation measures can catalyze community-wide progress. A good evaluation measure must quantify the most pertinent aspects of the system's performance while remaining reasonably simple and interpretable. Arriving at such a measure for navigation was a challenge because navigation can be quantified along several dimensions, including how close to the goal the agent got and how long it took to get there. The working group discussed a variety of factors that reflect navigation performance, as well as ways to combine these factors into summary measures. Based on these discussions, we make a number of recommendations.

The first issue we address is whether the agent needs to signal that is has completed the task. In some recent work, a navigation episode is terminated and deemed successful once the agent comes sufficiently close to the goal. We recommend against this, because such protocols do not test whether the agent understood that it has reached the goal. We consider such understanding essential: the agent must not merely stumble upon the goal, it must understand that the goal has been reached. To indicate such understanding, we recommend adding a dedicated action to the agent's vocabulary. The action can be called 'DONE' and indicates that the agent is ready to be evaluated. The agent's configuration relative to the goal, and the path it took to get there, should be evaluated when the agent produces this special signal, 'DONE'. Without such signal, a navigation episode should not be considered successful even if the agent came close to the goal.

**Recommendation 1.** The agent must be equipped with a special action that indicates that it has concluded the navigation episode and is ready to be evaluated. The agent's configuration relative to the goal must be evaluated at the time this action is produced, not at some favorable preceding time during the episode.

Our next recommendation concerns distance measures that should be used to evaluate proximity to the goal. We recommend against using Euclidean distance because it does not take the structure of the environment into account. For example, the agent may be close to the goal in Euclidean terms but be separated from it by a wall and in fact be far from it in terms of reachability: such configuration should not be considered close.

**Recommendation 2.** To measure proximity (for example, of the agent to the goal), we recommend using the geodesic distance, i.e., the shortest-path distance in the environment.

We now proceed to describe a specific measure that we recommend using to evaluate navigation performance. While a variety of factors can be informative and can be used as auxiliary indicators, we have sought a single summary measure that can be used as a primary quantifier in comparing the performance of different agents.

To define such a measure, we begin by adopting a binary criterion of whether a navigation episode is deemed successful. To evaluate this criterion, we consider the configuration of the agent at the time it outputs the action 'DONE'. (If no such action is produced by the agent, the episode is automatically deemed unsuccessful.) The episode is successful if at this time the agent is sufficiently close to the goal. For example, for PointGoal or ObjectGoal, the episode is successful if the distance between the agent and the goal is below a threshold $\tau$. We recommend that a threshold of $2\times$ the agent's body width is used by default. For AreaGoal, the episode is successful if the agent's center of mass is within the specified area.

Equipped with a binary definition of episodic success, we conduct $N$ test episodes. In each episode, the agent is tasked with navigating to a goal. Let $\ell_i$ be the shortest-path distance from the agent's starting position to the goal in episode $i$, and let $p_i$ be the length of the path actually taken by the agent in this episode. Let $S_i$ be a binary indicator of success in episode $i$. We define a summary measure of the agent's navigation performance across the test set as follows:

$$\frac{1}{N}\sum_{i=1}^{N} S_i \frac{\ell_i}{\max(p_i, \ell_i)}. \tag{1}$$

We will refer to this measure as SPL, short for Success weighted by (normalized inverse) Path Length. Let's consider some examples. If $50\%$ of the test episodes are successful and the agent takes the optimal path to the goal in all of them, its SPL is $0.5$. If all of the test episodes are successful but the agent takes twice as long to reach the goal as it could have if it had acted optimally, the SPL is also $0.5$. If $50\%$ of the test episodes are successful and $p_i = 2\ell_i$ in all of them, the SPL is $0.25$.

Note that SPL is a rather stringent measure. When evaluation is conducted in reasonably complex environments that have not been seen before, we expect an SPL of $0.5$ to be a good level of navigation performance. This can be calibrated more reliably on each dataset by measuring the SPL of human subjects.

**Recommendation 3.** We recommend adopting SPL as the primary measure of navigation performance.

While we recommend SPL as the primary evaluation measure, we note that there are other measures that can provide complementary information on the agent's performance. We encourage reporting such auxiliary measures in addition to SPL. Some suggestions for auxiliary measures are: a) success rate as a function of the normalized inverse distance traversed by the agent, b) distance to the goal at the end of the episode (absolute or normalized by $\ell_i$), c) sweep of SPL over varying thresholds $\tau$, d) distribution of the normalized inverse path length ($\ell_i / \max(p_i, \ell_i)$), e) number of infractions (e.g., contact with obstacles), and f) actuation time and energy spent in trajectory execution.

## 5. Experimental Testbeds

Much of the recent research on navigation has been conducted in simulation. Simulation has traditionally been regarded with skepticism in robotics. ("Simulations are doomed to succeed." [5]) This skepticism is well-taken. We must remember that embodied agents must eventually function in the physical world. However, simulation provides important advantages in terms of experimental methodology. Notably, simulation enables coordinated community-wide progress through reproducible experiments. Simulation supports the application of the common task framework [8] to embodied behavior. For this reason, we advocate broad adoption of simulation platforms by the community, accompanied by continued investment in techniques for transferring learned navigation skills from simulation to reality [26, 21].

A number of simulation platforms for indoor environments have recently been proposed [4, 12, 16, 28, 30, 31]. They are based on collections of environments, such as AI2-THOR [16], SUNCG [29], Matterport3D [7], and Gibson [31]. Other simulators that have been used to study navigation include ViZDoom and DeepMind Lab [15, 2] (relatively uncluttered corridor layouts), and CARLA [10] (outdoor urban environments). We refrain from recommending a particular simulator over others. Rather, we make two technical recommendations that can inform the design of future simulation platforms.

**Recommendation 4.** We recommend the use of continuous state spaces, such that the agent is free to move in continuous space. Discrete environments can be convenient and can support interesting experiments and results [1, 3, 19, 22], but continuous spaces better reflect the conditions in which agents will be deployed in the physical world.

**Recommendation 5.** For interpretability and interoperability, we recommend the adoption of SI units in simulators. Distance 1 in a simulator should correspond to 1 meter.

We believe customizability of these simulation platforms will facilitate exploration of new problems and novel ideas, as well as controlled evaluation. However, this customizability comes with responsibility. It is expected that research that customizes these simulators (e.g., to study specific aspects of a standard scenario, to study a new problem, or to study problems with modified sensors or actuators) will properly document the customizations made to the simulators by releasing open-source code as well as clearly stating the underlying problem assumptions (such as extent of actuation noise or availability of accurate odometry) and agent configurations (such as specifics of the observation, state, and control space). Unlike physical robotics settings, where it is typically impossible to match the exact setup of past experiments, we hope that research performed using simulation platforms will provide materials necessary to reproduce the precise experimental conditions and results.

To ensure that research conducted in simulation does not lose its connection to physical systems, it is imperative that research findings obtained in simulation can be transferred and deployed on physical robots. Therefore, we recommend that simulators be accompanied by open-source software that supports the deployment of agents trained in simulation to real-world physical platforms. For example, such software could consist of ROS packages [24] exposing an API identical to an associated simulation platform, possibly including a domain adaptation module such as 'goggles' [31]. Initially, these attempts may expose various limitations and obstacles to real-world deployment that may not have been fully considered during simulator design. However, over time, standard deployment frameworks can provide a direct pathway for simulated successes to translate into real-world impact.

**Recommendation 6.** We recommend that simulators are supported by open-source software enabling agents trained in simulation to be deployed to physical robots using standard components.

## 6. Standard Scenarios

To establish a benchmark for evaluating the navigation performance of embodied agents we provide a set of standard scenarios. We share these standard scenarios publicly to promote consistency in evaluation and to enable direct comparison.

We create these standard scenarios by first selecting subsets of scenes from each of the currently available environment datasets (SUNCG, AI2-THOR, Matterport3D, Gibson). For each dataset, we select a subset of scenes and split them into training, validation, and test sets.

A scenario specifies the starting state of the agent and its goal state. We sample starting and goal positions at random from points in the free space of each environment. Each generated combination of start and goal positions is checked for navigability using a tile-based shortest-path computation. We generate more than 100 configurations per scene, spanning a range of geodesic distances to the goal from a minimum of 1 m to an unbounded maximum determined by the size of the specific scene. We now describe the details of the scenarios for each dataset.

**SUNCG.** We use 500 single-floor SUNCG houses of varying complexity (with one to ten rooms per house), split into 300/100/100 training/validation/test environments. These houses consist of a total of 2,737 rooms, populated with 41,158 objects in a total floor area of approximately 110,000 $m^2$. On average, there are 5.5 rooms per house with a mean floor area of 42 $m^2$ per room. These environments represent a variety of interiors including family homes, offices, and public spaces such as restaurants. The agent has a body-width of 0.2 m and we set the distance threshold $\tau$ for successful navigation to the goal to twice the body-width (0.4 m). The scenario specification files are available at github.com/minosworld/scenarios.

**Matterport3D.** We adopt the 61/11/18 training/validation/test house splits specified by the original dataset [7]. These houses comprise a total of 2,206 room regions in 190 floors. On average, each house has 24.5 rooms and a floor area of 560 $m^2$. These environments represent mostly private residences, hotels, and company office spaces. Similarly to the SUNCG scenarios, the agent body-width is 0.2 m and the distance threshold $\tau$ is 0.4 m. The scenario specification files are available at github.com/minosworld/scenarios.

**AI2-THOR.** Version 1.0 of AI2-THOR includes 120 scenes covering four scene categories: kitchens, living rooms, bedrooms, and bathrooms. The first 20 scenes in each category should be used for training (e.g., FloorPlan1 to FloorPlan20 for kitchens). The next five scenes in each category should be used for validation (e.g., FloorPlan21 to FloorPlan25 for kitchens), and the last five scenes for testing. The training should be performed on all training scenes (80 scenes in total). We consider two types of goal specification for AI2-THOR: PointGoal and ObjectGoal. The current version of AI2-THOR is not suited for AreaGoal since each scene consists of only one room. Below we describe the specifications for each goal setting.

**PointGoal:** For each test scene, we randomly choose 5 points as the target and provide the target coordinates to the agent. We also consider 5 different starting points for the agents and 5 different scene configurations (by moving objects to different locations or changing their state). In total, we have $5 \times 5 \times 5 = 125$ scenarios for each test scene. The episode is considered successful if the distance between the agent and the target point is below $2\times$ the agent's width. The action set for this setting is: *Move Forward*, *Move Backward*, *Rotate Right*, *Rotate Left*, and *Terminate*.

**ObjectGoal:** We provide the object category label to specify a target. The navigation is successful if the object is 'visible' to the agent. In AI2-THOR, an object is marked as 'visible' if it is within 1 m from the camera and is in the field of view of the agent. Similar to PointGoal, we consider 5 different random starting points for each test scene. There are two scenarios for choosing the targets:

1. *Navigation-only:* In this setting, the targets are chosen in a way that can be reached only by navigation actions. For example, a mug inside a cabinet cannot be reached by navigation actions alone since the agent needs to open the cabinet to see the mug. The action set for this setting is: *Move forward*, *Move backward*, *Rotate Right*, *Rotate Left*, *Look Up*, *Look Down*, and *Terminate*.

2. *Interaction-based:* In this setting, finding some of the targets requires interaction. The action set for this setting is: *Move Forward*, *Move Backward*, *Rotate Right*, *Rotate Left*, *Look Up*, *Look Down*, *Open X*, *Close X*, and *Terminate*.

Similar to the PointGoal setting, we consider 5 object categories and 5 scene configurations for each test scene (125 scenarios per scene).

**Gibson.** The Gibson dataset includes 572 buildings composed of 1,447 floors covering a total area of 211,000 $m^2$. The spaces are real buildings scanned using 3D scanners. Each building comes with a measure of clutter (SSA) and navigation complexity. Visualization of the spaces and their metadata is available at http://gibson.vision/database/. Given the size of the dataset, we specify a few different standard partitions with various sizes to facilitate experimentation, such as tiny (35 buildings), medium (140 buildings), and full (572 buildings). The train/validation/test splits for each partition are: tiny (25/5/5), medium (100/20/20), full (402/85/85), all roughly 70%/15%/15%.

For each space, 100 pairs of points $(initial\ location, target\ location)$ are provided as the fixed standard scenarios. The locations are picked randomly and are verified to be reachable (defined as: having 1 $m$ of open space around the points and not being over/under other objects, e.g. over shelves, under desks). The minimum distance between initial and target locations is 1 $m$, and the maximum depends on the size and navigation complexity of each space. The scenario files can be accessed at http://gibson.vision/nav_scenarios/.

The agents in the Gibson simulator can be specified by the user, and thus the action space is not fixed. The simulator provides optional controllers that can be used to abstract the action space into discrete (e.g. forward, backward, left, right) or continuous (e.g. velocity, angle) navigational commands. Different agents also incur different geodesic distances (e.g., a ground vehicle has to travel on the floor while a drone can fly over furniture). We computed the default geodesic distance between points at a fixed height over the floor, though it can be recalculated for each particular agent. The initial and target points are picked from the same floor to remove the need to climb stairways as the feasibility/complexity of that becomes agent-specific. We suggest $2\times$ the agent's body width as the distance threshold $\tau$ for successful navigation, with $0.4\ m$ as the default value.

## 7. Agent Architectures

The simplest navigation agents are purely reactive. In each time step, sensory input from the environment is processed by a deep network that outputs an action. The agent does not carry internal state across time steps and does not construct an internal representation of its environment [9, 32]. A small departure from purely reactive architectures is to equip the agent with short-term vectorial memory maintained by a recurrent update mechanism [14, 17, 20]. The next level of architectural sophistication is to use more advanced memory mechanisms that support the construction of rich internal representations of the agent's environment [12, 13, 22, 23, 27]. We view the nature of the internal representation maintained by the agent as central to the study of embodied navigation. It is much too early to recommend any particular form of representation. We merely highlight the importance of this line of inquiry.

**Recommendation 7.** A central issue in the design of embodied agents is the structure of the internal representation that the agent constructs and maintains as it navigates through its environment. We encourage comprehensive and open-minded study of this question, which we view as fundamental to the development of Artificial Intelligence.

## 8. Acknowledgments

We thank the following for detailed and constructive feedback on an earlier draft of this report: Samarth Brahmbhatt, Andrew Davison, Aleksandra Faust, David Fouhey, Raia Hadsell, James Hays, Derek Hoiem, Philipp Krähenbühl, John Leonard, Sergey Levine, Benjamin Recht, Stephan Richter, Mathieu Salzmann, Konrad Schindler, Fei Sha, Josef Sivic, Anton Van Den Hengel, Qi Wu.